# GaborNet: Gabor filters with learnable parameters in deep convolutional neural networks

**Andrey Alekseev**
Moscow Institute of Physics and Technology
`alekseev.as@phystech.edu`

**Anatoly Bobe**
MIPT Neurorobotics lab.
`anatoly_bobe@mail.ru`

MIPT Neurorobotics lab., Institutsky per. 9/7, Moscow Region, Dolgoprudny 141700, Russian Federation

## Abstract

The article describes a system for image recognition using deep convolutional neural networks. Modified network architecture is proposed that focuses on improving convergence and reducing training complexity. The filters in the first layer of the network are constrained to fit the Gabor function. The parameters of Gabor functions are learnable and are updated by standard backpropagation techniques. The system was implemented on Python, tested on several datasets and outperformed the common convolutional networks.

## Introduction

Convolutional neural networks (CNNs) have various applications, especially in computer vision. They have attracted significant attention due to the ability to be trained end-to-end and capability of learning outstanding feature representations from raw image data. In 2015 CNN showed human-like performance on ImageNet dataset [1]. Unlike classic computer vision methods, a neural network is a data-driven algorithm that learns robust representations from data, but usually at the cost of training an excessive number of parameters (or weights). Additionally, a convergence of the neural network depends on parameters initialization. Usually, the weights are initialized by uniform or normal distribution. But it causes the convergence problem and makes it difficult to train very deep CNN [2].

In [2] Glorot and Bengio proposed a formula that estimates standard deviation for a CNN layer based on layer input and output size. In [1] new method of initialization, known as He initialization, was introduced, which also suggests using normal or uniform distribution with parameters based on neural network topology and type of activation functions. However, these approaches focused mainly on dealing with vanishing gradient problem while training deep CNNs. Hence additional constraints and initialization methods of some other nature could be applied to convolutional neural networks' weights to improve the convergence process. Specifically, this could be achieved by generating weights of a neural network by a family of parameterized filters.

Gabor filters [3] are widely used in computer vision. They are based on a sinusoidal plane wave with particular frequency and orientation, which allows them to extract spatial frequency structures from images [4]. Additionally, these filters proved to be appropriate for texture representation and face detection tasks [5]. A few works have explored Gabor filters for CNN. In [6] Gabor filters were used as preprocessing tool to generate Gabor features then using it as an input to a CNN, in [7] first or second layer of CNN was set as a constant Gabor filter bank, thus reducing number of trainable parameters of the network, and in [8] Convolutional Gabor orientation Filters were introduced, a special structure that modulates convolutional layers with learnable parameters by non-learnable Gabor filter bank. However, the authors did not report the integration of the filter parameters into the backpropagation algorithm.

In this paper, we propose using Gabor Layer as the first layer in a deep convolutional network. Gabor Layer is a convolutional layer, which filters are constrained to fit Gabor functions (Gabor filters). The parameters of filters are initialized from filter bank, proposed in [9], and during the training process they are updated by the standard backpropagation algorithm. This approach aims to increase the robustness of learned feature representations and to reduce training complexity of neural networks. Gabor Layer is implemented on basic elements of CNNs and can be easily integrated into any deep CNN architecture.

## Theory

### CNN

Convolutional neural networks are usually presented as a sequence of several convolutional layers and fully connected layers, with nonlinear activation function applied to the output of each layer. Pooling and dropout layers may be included in the architecture to avoid overfitting. Convolutional layers reduce memory usage and increase performance by using the same filters for each pixel of the image. The standard architecture of CNN is shown in Figure 1.

The training process of CNN is usually carried out using backpropagation algorithm [10] which iteratively collects the gradient values for the weight coefficients located on different layers. The weight update is then performed using different variations of stochastic gradient descent (SGD) methods. Commonly a family of first-order adaptive SGD is utilized [11] [12]. Thus imposing restrictions on convolutional layer generating function, such as monotony and differentiation, is required.

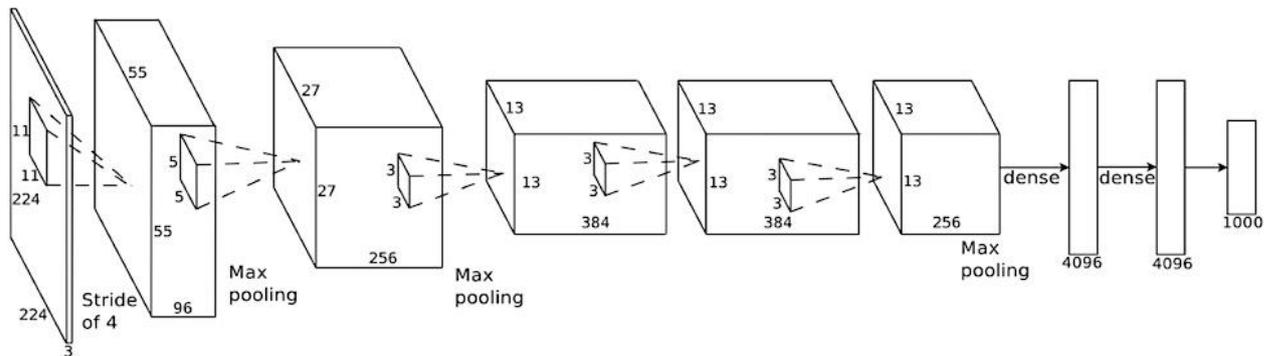

*Figure 1 CNN structure (Alexnet) from http://ml4a.github.io/*

### Gabor function

Gabor function is a complex sinusoid modulated by Gaussian, which satisfies the requirements of monotonicity and differentiability, i.e.

$$g(x, y, \omega, \theta, \psi, \sigma) = exp\left(-\frac{x'^2 + y'^2}{2\sigma^2}\right) exp(i(\omega x' + \psi)) \quad (1)$$

$$x' = x \cos\theta + y \sin\theta \quad (2)$$

$$y' = -x \cos\theta + y \cos\theta \quad (3)$$

However, in this work, we used real values of Gabor function i.e.

$$g(x, y, \omega, \theta, \psi, \sigma) = exp\left(-\frac{x'^2 + y'^2}{2\sigma^2}\right) cos(\omega x' + \psi) \quad (4)$$

Gabor filters proved to be an efficient tool for extracting spatially localized spectral features, which are used in various pattern analysis applications [5]. In [11] was shown that deep CNN trained on real-life images tends to learn first convolutional layers contain mostly Gabor-like filters. Filters of the first layers of AlexNet is shown in Figure 2. This corroborates an idea of using Gabor filters in the first layer of CNN.

In real image processing applications, the features are usually extracted using a bank of Gabor filters, which parameters are set according to the approach proposed in [12]. Frequencies $\omega_n$ and orientations $\theta_m$ of the Gabor filters are obtained by the following equations:

$$\omega_n = \frac{\pi}{2}\sqrt{2}^{-(n-1)} \quad (5)$$

$$\theta_m = \frac{\pi}{8}(m-1) \quad (6)$$

$$n = 1, 2, \dots, 5$$

$$m = 1, 2, \dots, 8$$

The $\sigma$ is set by $\sigma \approx \pi/\omega$, which allows to define the relationship between $\sigma$ and $\omega$. The $\psi$ is set by uniform distribution U(0, $\pi$). In our work, we initialized the Gabor Layer weights the same way.

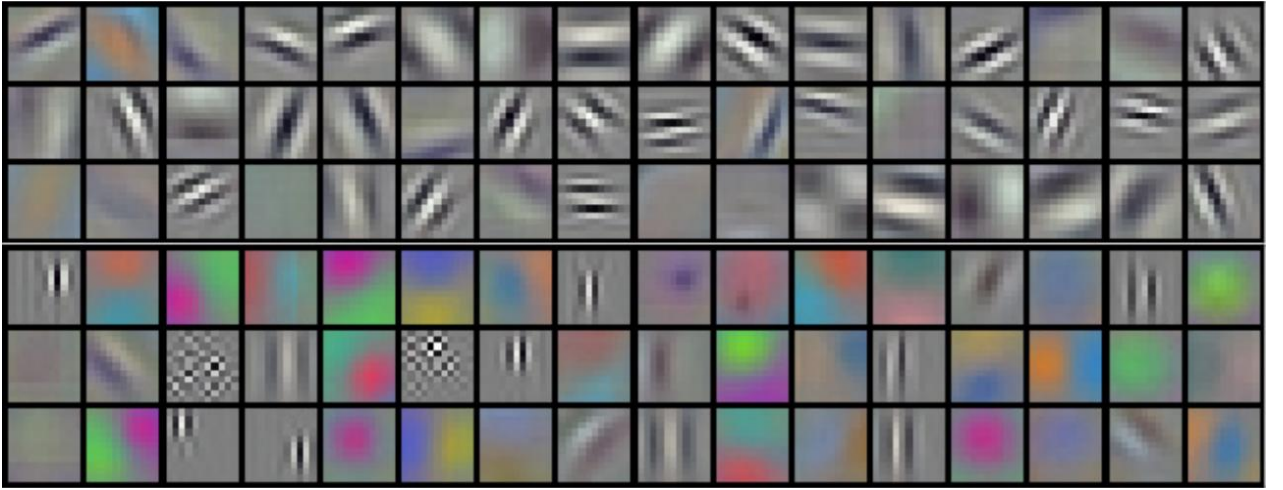

*Figure 2  96 convolutional kernels of size 11×11×3 learned by the first convolutional layer on the 224×224×3 input images in Alexnet*

## Experiments

In this section, we present the performance details of different CNN architectures explored during the experiments. Every CNN architecture was implemented in two ways: regular CNN and CNN with Gabor Layer as the first layer of the network (Gabor CNN or GCNN) in order to estimate the effect of Gabor layer on convergence speed and performance metrics. During the experiments the following datasets were used: Dogs vs Cats dataset [13], human emotions dataset AffectNet [14] and IMAGENET ILSVRC2012 [15].

## Dogs vs Cats

This dataset consists of 25000 colored images of different size. For validation purposes 30% of the dataset was used. All experiments were performed with a batch size of 64. Adam [16] was used as an optimization algorithm for both CNN and GCNN architectures. Pooling [17] and dropout [18] layers, as well as ReLU [19] activation function, were also used. Images were normalized. Figure 3 shows the architecture of CNN and GCNN.

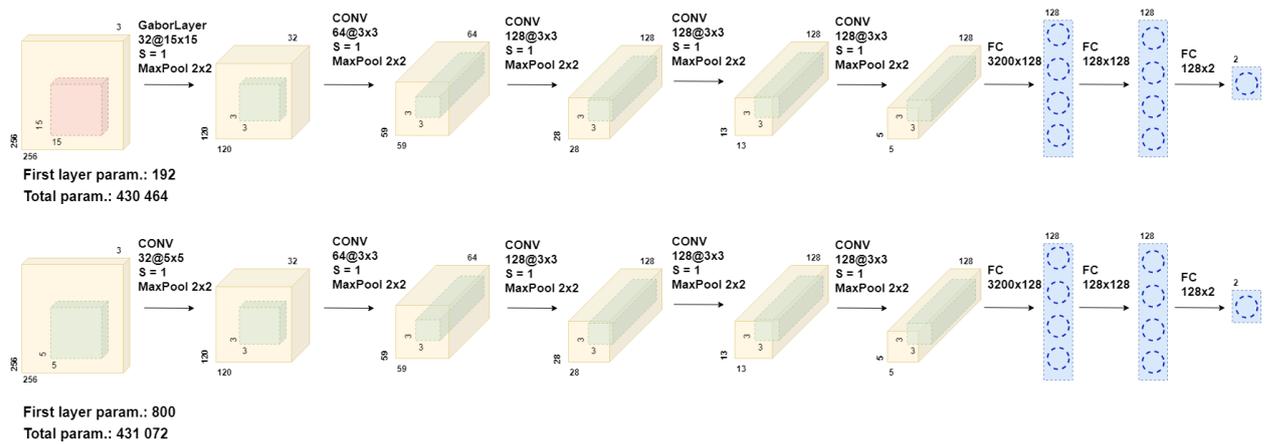

*Figure 3. The architecture of GCNN and CNN used for Dogs vs Cats dataset*

Both networks were trained for 100 epochs. Learning rate was set to 0.001 and betas = (0.9, 0.999). Results are shown in Figure 4. As it can be seen, GCNN outperforms regular CNN and converged on earlier epochs. The performance gap achieves 6% accuracy in the end. The performance of GCNN and CNN is listed in detail in Table 1.

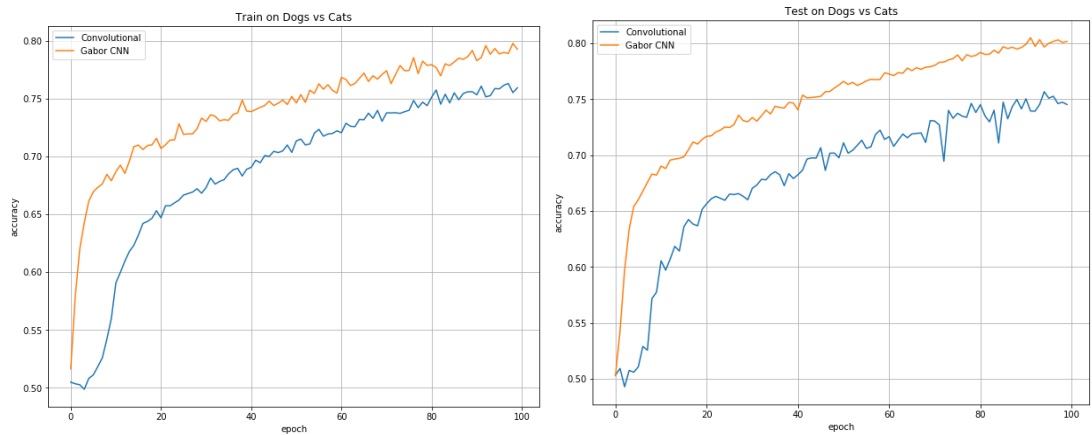

*Figure 4. The performance of GCNN and CNN on Dogs vs Cats dataset*

*Table 1. Accuracy score on Dogs vs Cats dataset*

|       | Train |           | Test  |           |
|-------|-------|-----------|-------|-----------|
| Epoch | CNN   | Gabor CNN | CNN   | Gabor CNN |
| 1     | 0.506 | 0.503     | 0.503 | 0.517     |
| 3     | 0.520 | 0.597     | 0.515 | 0.620     |
| 10    | 0.613 | 0.682     | 0.616 | 0.679     |
| 40    | 0.674 | 0.747     | 0.668 | 0.739     |
| 90    | 0.732 | 0.796     | 0.726 | 0.792     |

# AffectNet

The AffectNet dataset consists of 420 299 colored images of human emotions filmed in the wild. For evaluation 250 000 images of 5 classes (Neutral, Happy, Sad, Surprise, Anger) were used. The structures of CNN and Gabor CNN are shown in Figure 5.

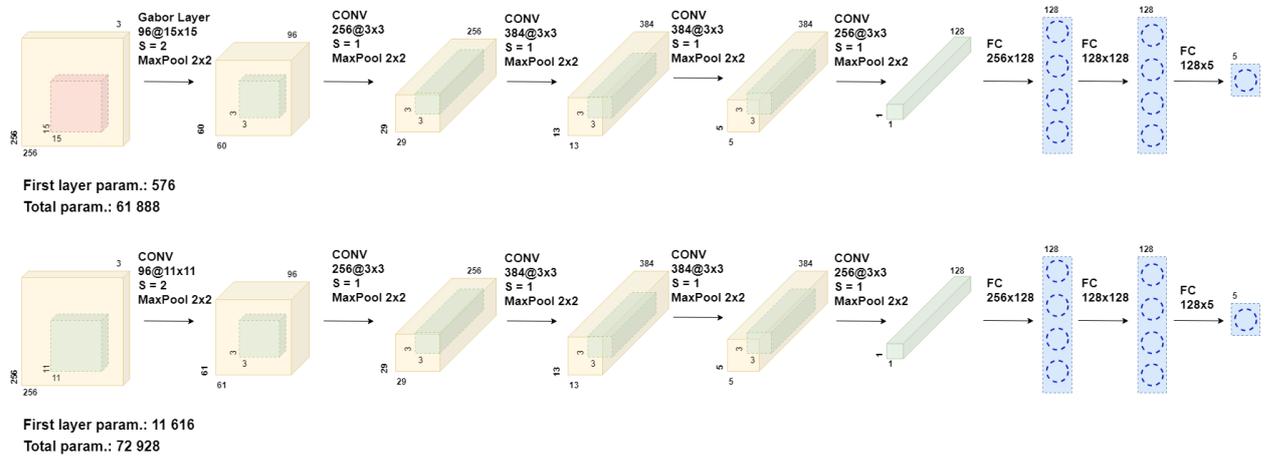

*Figure 5. The architecture of GCNN and CNN used for AffectNet dataset*

Gabor CNN achieves better results on earlier epochs both on a train and on a test. The average difference is 3% of the accuracy score. In addition, Gabor CNN converges several epochs earlier than regular CNN. However, unlike Dogs vs Cats dataset, on later epochs, CNN achieves almost the same accuracy score as Gabor CNN and the average accuracy[1] difference drops to 1%. The performance gap is shown in Figure 6. The performance of GCNN and CNN is listed in detail in Table 2.

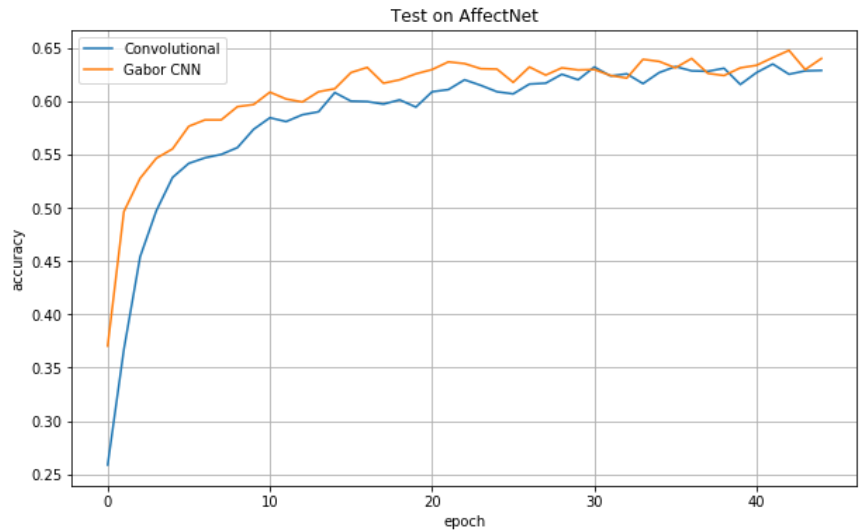

---

[1] Due to oscillating values of accuracy score, moving average with window size of 5 epochs was used. The plots of moving average for testset is in the appendix.

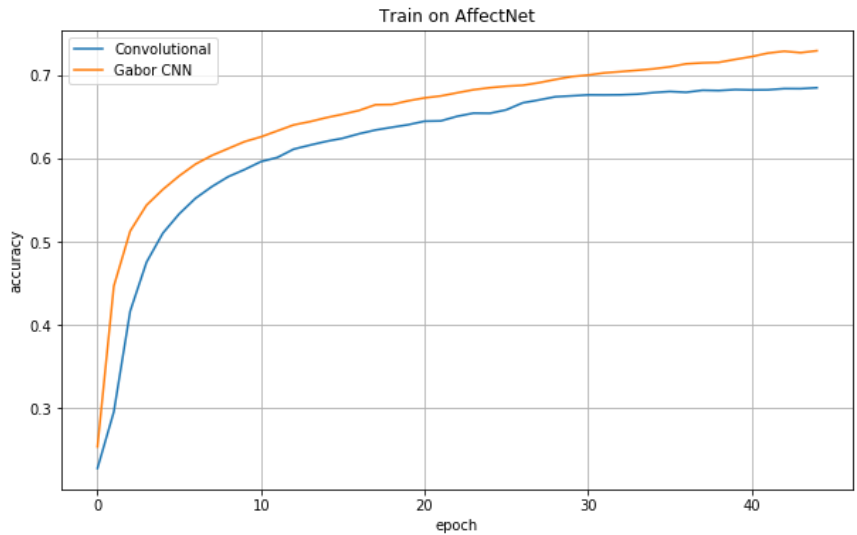

*Figure 6 The performance of CNN and Gabor CNN*

*Table 2 Accuracy score on AffectNet dataset*

|  | Train |  | Test |  |
|---|---|---|---|---|
| Epoch | CNN | Gabor CNN | CNN | Gabor CNN |
| 1 | 0.228 | 0.254 | 0.256 | 0.370 |
| 5 | 0.510 | 0.563 | 0.528 | 0.555 |
| 10 | 0.586 | 0.620 | 0.574 | 0.597 |
| 20 | 0.640 | 0.669 | 0.594 | 0.626 |
| 35 | 0.679 | 0.710 | 0.627 | 0.637 |
| 45 | 0.685 | 0.719 | 0.629 | 0.640 |

## IMAGENET

For evaluation on IMAGENET, an unchanged dataset from ILSVRC 2012 was used. Data augmentation techniques, such as random horizontal flip and random crop, proposed in [11] were used. Here the original Alexnet architecture was used [11]. The learning rate was decreased by factor 10 on $30^{th}$ and $50^{th}$ epochs. Unlike on Dogs vs Cats and AffectNet datasets, the network with Gabor Layer had smaller accuracy score on earlier epochs on IMAGENET. However, from 10 to 30 epochs the performance gap between Gabor Alexnet and Alexnet reached almost 2% on top1 and top5 scores in favor of Gabor Alexnet. After reducing the learning rate on $30^{th}$ epoch Alexnet leveled top1 and top5 scores with Gabor Alexnet. By the end of the training stage, Gabor Alexnet and Alexnet achieved almost exact results. The performance of Gabor Alexnet and Alexnet is shown in Figure 7. The performance of Gabor Alexnet and Alexnet is listed in detail in Table 3.

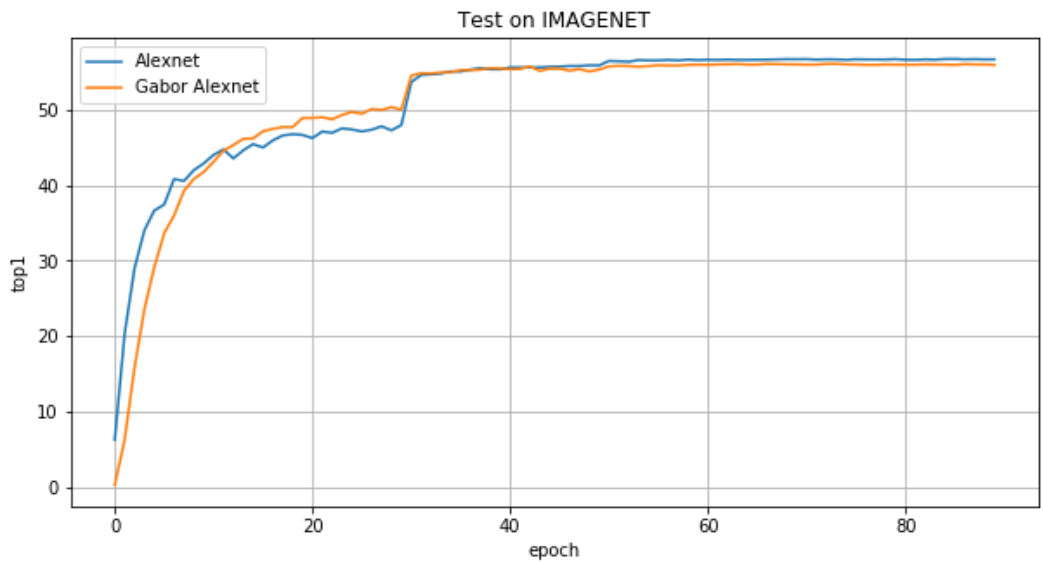

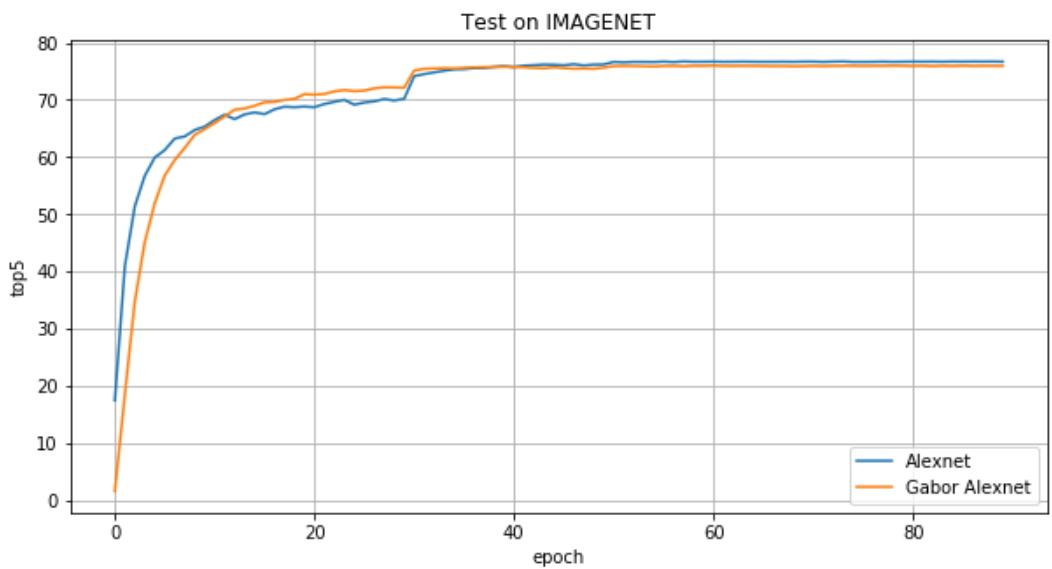

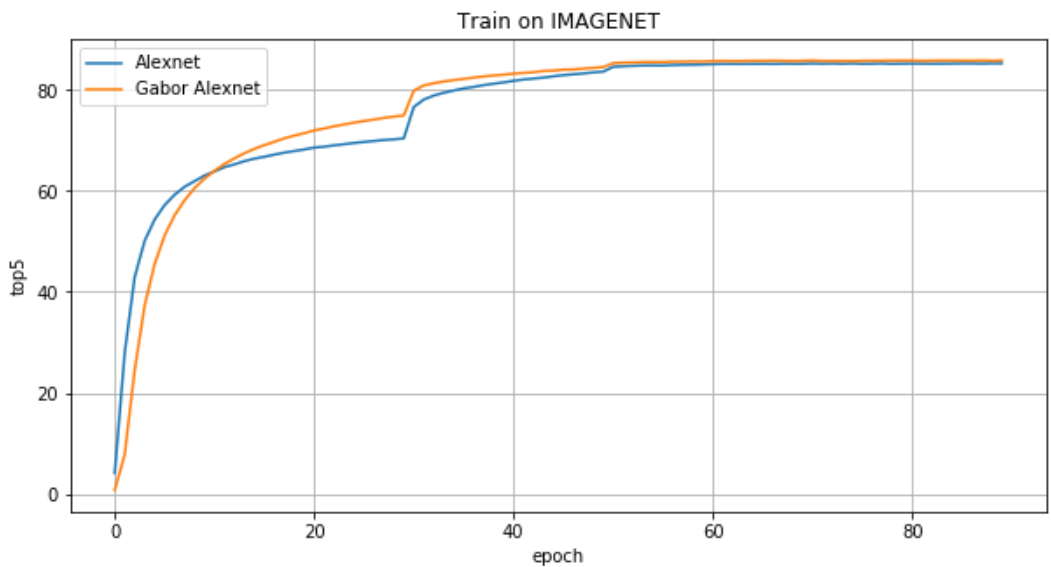

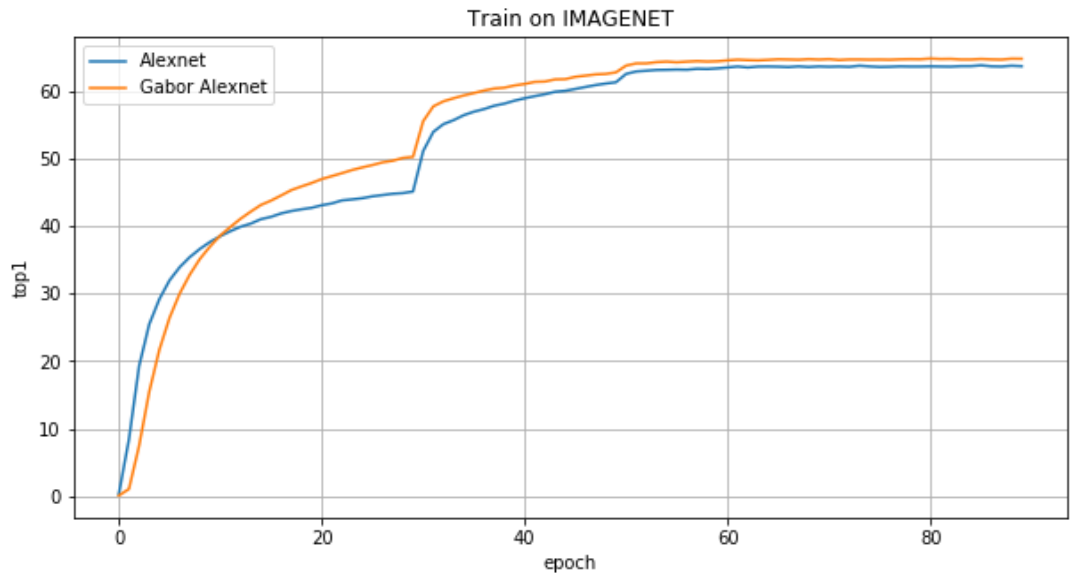

*Figure 7. The performance of Alexnet and Gabor Alexnet*

*Table 3. The accuracy score in percent*

|  | Top1 | | | | Top5 | | | |
|---|---|---|---|---|---|---|---|---|
|  | Train | | Test | | Train | | Test | |
| Epoch | Alexnet | Gabor Alexnet | Alexnet | Gabor Alexnet | Alexnet | Gabor Alexnet | Alexnet | Gabor Alexnet |
| 1 | 0.32 | 0.11 | 6.28 | 0.29 | 4.13 | 0.75 | 17.52 | 1.67 |
| 5 | 29.18 | 21.71 | 36.65 | 29.15 | 54.33 | 45.50 | 59.93 | 51.95 |
| 10 | 37.64 | 37.00 | 42.90 | 41.75 | 63.04 | 62.40 | 65.33 | 64.87 |
| 15 | 41.00 | 43.10 | 45.45 | 46.21 | 66.37 | 68.32 | 67.81 | 68.97 |
| 25 | 44.10 | 48.66 | 47.40 | **49.71** | 69.49 | 73.53 | 69.16 | **71.53** |
| 45 | 59.98 | 61.71 | 55.70 | 55.45 | 82.70 | 83.81 | 76.14 | 75.70 |
| 70 | 63.60 | 64.62 | 56.71 | 56.10 | 85.19 | 85.76 | 76.68 | 75.90 |
| 90 | 63.63 | 64.73 | 56.68 | 56.01 | 85.27 | 85.83 | 76.70 | 76.00 |

## Discussion and future work

CNNs with Gabor Layers show better performance on several datasets. CNN with utilizing Gabor Layer on «Dogs vs Cat» dataset significantly outperforms «classic» CCN up to 6% in accuracy score. It is due to a better first approximation of Gabor filters on real-life images and better convergence as the number of learnable parameters is reduced up to 21 times. However, if images in datasets contain less Gabor-like features, this approach will not improve accuracy score of the neural network, as shown in the experiment on IMAGENET. The training time for one epoch of GCNN takes 1.2-2 more time than training time of classic CNN because of the difficulty of calculating of Gabor filters. However, use of GCNN is justified due to its ability to achieve the same accuracy score as classic CNN but 1.4-4 times faster depending on the dataset. The future work will focus on using different filter banks, utilizing deeper layers of the neural network and adapting this algorithm as a preprocessing tool.

# Appendix

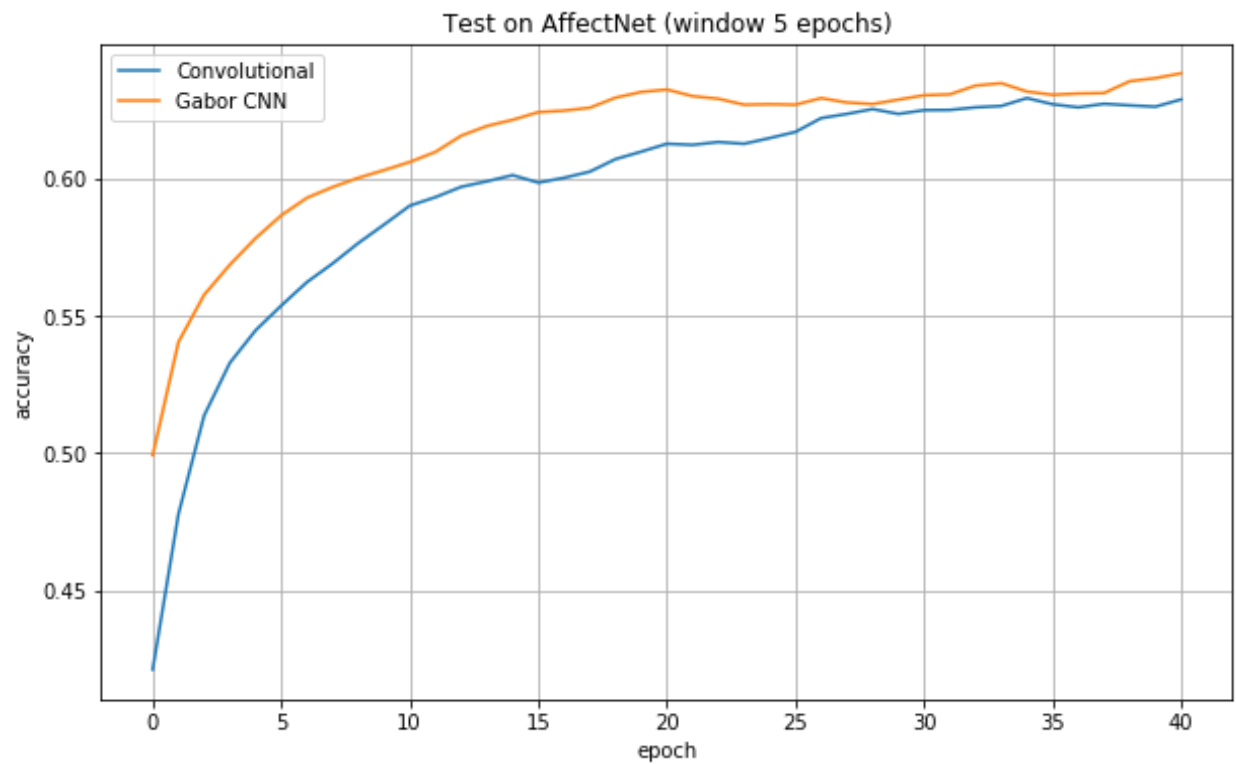